# Multi-Ontology Refined Embeddings (MORE): A Hybrid Multi-Ontology and Corpus-based Semantic Representation for Biomedical Concepts

Steven Jiang[1], Weiyi Wu[2], Naofumi Tomita[2], Craig Ganoe[2], Saeed Hassanpour[1,2,3*]

[1]Department of Computer Science, Dartmouth College, Hanover, NH 03755

[2]Department of Biomedical Data Science, Geisel School of Medicine at Dartmouth, Hanover, NH 03755

[3]Department of Epidemiology, Geisel School of Medicine at Dartmouth, Hanover, NH 03755, USA

*Corresponding Author: Saeed Hassanpour, PhD

Postal address: One Medical Center Drive, HB 7261, Lebanon, NH 03756, USA

Telephone: (603) 650-1983

Email: Saeed.Hassanpour@dartmouth.edu

# ABSTRACT

**Objective:** Currently, a major limitation for natural language processing (NLP) analyses in clinical applications is that a concept can be referenced in various forms across different texts. This paper introduces Multi-Ontology Refined Embeddings (MORE), a novel hybrid framework for incorporating domain knowledge from multiple ontologies into a distributional semantic model, learned from a corpus of clinical text.

**Materials and Methods:** We use the RadCore and MIMIC-III free-text datasets for the corpus-based component of MORE. For the ontology-based part, we use the Medical Subject Headings (MeSH) ontology and three state-of-the-art ontology-based similarity measures. In our approach, we propose a new learning objective, modified from the Sigmoid cross-entropy objective function.

**Results and Discussion:** We evaluate the quality of the generated word embeddings using two established datasets of semantic similarities among biomedical concept pairs. On the first dataset with 29 concept pairs, with the similarity scores established by physicians and medical coders, MORE's similarity scores have the highest combined correlation (0.633), which is 5.0% higher than that of the baseline model and 12.4% higher than that of the best ontology-based similarity measure. On the second dataset with 449 concept pairs, MORE's similarity scores have a correlation of 0.481, with the average of four medical residents' similarity ratings, and that outperforms the skip-gram model by 8.1% and the best ontology measure by 6.9%.

**Conclusion:** MORE incorporates knowledge from several biomedical ontologies into an existing corpus-based distributional semantics model (i.e., word2vec), improving both the accuracy of the learned word embeddings and the extensibility of the model to a broader range of biomedical concepts. MORE allows for more accurate clustering of concepts across a wide range of applications, such as analyzing patient health records to identify subjects with similar pathologies or integrating heterogeneous clinical data to improve interoperability between hospitals.

# INTRODUCTION

With the increasing availability of health-related textual data, such as Electronic Health Records (EHR), novel applications of Natural Language Processing (NLP) in the field of medical informatics are a growing topic of interest [1–7]. Currently, a major limitation of NLP analysis techniques for clinical text is related to the free-text format of these records and notes—the same concept can be referenced in various forms across different texts (e.g., "kidney failure" and "renal failure"). In particular, different physicians and institutions may use unique terminologies for reporting the same concepts in EHRs. To address this issue, researchers use semantic similarity measures to identify similar biomedical concepts in free-text records and notes. A semantic similarity measure takes two concepts as input and returns a numeric score that quantifies how alike they are in meaning [1].

A hybrid biomedical semantic similarity measure can improve the identification and clustering of biomedical concepts across a wide range of applications, improving patient care and clinical outcomes [2,8]. For example, patient health records can be analyzed to identify subjects with similar conditions or pathologies. With this information, data-mining techniques can be used to extract useful information about previous care processes, the evolution of certain diseases, and social trends [9]. Semantic similarity measures can also assist in identifying patients for clinical studies and clustering symptoms in clinical text for post-marketing medication safety surveillance [8]. Furthermore, they can be used to integrate heterogeneous clinical data, which can improve interoperability between medical sources and allow hospitals to share patient health information more effectively [9]. Finally, in the fields of medical information retrieval and literature mining, users' queries can be extended to conceptually equivalent formulations to improve keyword-based search engines [9]. Ultimately, semantic similarity measures can improve the statistical power of NLP analyses [10], making it easier to identify associations between conditions and clinical outcomes in health records and improve information retrieval from scientific journals and clinical reports [8].

A variety of semantic similarity measures have been developed to describe the strength of the relationships between concepts in biomedicine. These existing semantic similarity measures mostly fall into two common categories: *ontology-based* or *corpus-based* semantic similarities. Ontology-based semantic similarities typically rely on different graph-based features [2,9], such as the shortest path length between concepts and the position of their lowest common ancestors, to capture semantic similarity. These ontology-based approaches depend on the completeness and quality of the underlying ontologies [9]; however, curating and maintaining domain ontologies is a labor-intensive and complicated task.

As an alternative to ontology-based semantic similarity, corpus-based semantic similarities are based on distributional semantics and co-occurrences of terms in the free text [11,12]. These corpus-based models rely on the linguistic principle that the meaning of a word (i.e., semantics) can be inferred based on its surrounding words (i.e., context). With recent advances in deep learning and the widespread use of distributional semantics to construct word embeddings for word representation in deep-neural networks, these corpus-based models have gained vast popularity. The word2vec [13] distributional semantics model is the most common method for generating such word embeddings. Intuitively, the word2vec model is a neural network that maps words with similar context to nearby points in a vector space. The cosine similarity between resulting word representations is commonly considered to be a corpus-based semantic similarity in various settings [14–16]. Although corpus-based semantic similarities are generated by unsupervised models, making them more extensible to a broader range of concepts, the lack of human curation and availability of relevant biomedical corpora limit their accuracy and usability in biomedical applications [2,16].

Previous efforts to combine ontology-based and corpus-based similarities to better capture semantic similarities between biomedical concepts [17–19] are hybrid approaches that mostly rely on the frequency of the appearance of those concepts in a corpus to compute information content, rather than considering the free-text context for measuring similarities. Additionally, existing hybrid measures for semantic similarity in biomedical domains do not incorporate ontological knowledge into the process of generating word embeddings. In non-biomedical domains, a few approaches have been developed to use constraints among words, such as word categories [20], links [21,22], or typed relations [20,22,23], as regularization terms in model training to construct better word embeddings.

In this paper, we propose Multi-Ontology Refined Embeddings (MORE) semantic similarity to effectively integrate ontological knowledge and corpus-based context into a novel semantic similarity measure. MORE uses existing ontology-based semantic similarity measures from the Unified Medical Language System (UMLS) to modify the objective function of the word2vec skip-gram model, a popular distributional semantic model. In our approach, we propose a mathematical framework for vector representation refinement that relies on a collection of the most established and reliable ontology-based measures, rather than a single ontology-based similarity, to maximize the utility of our measure in a broad domain. In other words, MORE uses multiple ontology-based semantic similarities as the overall indicator of ontological similarity to refine the distributional semantic representations. Of note, our implementation is based on the official TensorFlow implementation of word2vec, and we have made it available for public use.[1] Our model is benchmarked against existing state-of-the-art semantic similarities using an established evaluation dataset for the semantic similarity between biomedical concepts. We find that MORE outperforms the baseline corpus-based semantic similarity model, as well as the individual ontology-based semantic similarities, in terms of correlation with physician and medical coder similarity scores in the evaluation dataset. The main contributions of this paper are two-fold: 1) we present a generalizable and extensible framework for incorporating domain-specific knowledge into a distributional semantic model, and 2) we show that this hybrid framework outperforms the baseline word2vec model and ontology similarity measures on two established benchmarks. In the remainder of this paper, we provide context for corpus-based, ontology-based, and hybrid semantic similarity measures in the biomedical domain. We also discuss the following components: the corpora used to train the corpus-based component of the model (i.e., RadCore and MIMIC-III), the UMLS-Similarity ontology measures used to modify the objective function, the mathematical framework used for modifying the cross-entropy objective function, and the benchmark dataset against which the proposed method was evaluated. We also discuss the results from evaluating the proposed measure against state-of-the-art benchmarks, present a conclusion, and propose directions for future research.

---

[1] https://github.com/BMIRDS/MORE

# BACKGROUND AND RELATED WORK

**Ontology-based Methods**

In the biomedical domain, there are a growing number of ontologies or hierarchical knowledge bases that represent semantic relationships between concepts. One of the best examples of this is UMLS, which is maintained by the National Library of Medicine (NLM) and includes two of the largest and most extensive ontology knowledge bases: Systematized Nomenclature of Medicine-Clinical Terms (SNOMED-CT) and Medical Subject Headings (MeSH) [8]. Ontology-based semantic similarity measures are based on "is-a" relations found in the underlying taxonomy or ontology in which the concepts reside. For example, the terms "common cold" and "illness" are similar because "common cold is a kind of illness. Likewise, common cold and influenza are similar in that they are both kinds of illness" [1]. As these ontology-based approaches are sensitive to the completeness and quality of the underlying ontologies [9], curating and maintaining domain ontologies is critical to guarantee the accuracy and robustness of ontology-based semantic similarities. Although there have been major efforts, such as the ongoing support by NLM, to curate and maintain biomedical ontologies as valuable sources of domain knowledge, it is a labor-intensive and elaborate task. Furthermore, due to the heterogeneity of biomedical domains and their corresponding concepts, there is no single top-performing ontology-based similarity measure across all domains and applications [9,24].

**Corpus-based Methods**

In 2013, Mikolov et al. [25] introduced the word2vec distributional semantics model, a neural network that maps words with similar context to nearby points in a vector space. As previously noted, the semantic similarity between two words is calculated as the cosine similarity of the generated vectors representing those words, ranging from -1 to 1. The paper of Mikolov et al introduced two model architectures: Continuous Bag-of-Words (CBOW) and skip-gram. While these models are algorithmically similar, CBOW predicts target words from context words, whereas skip-gram predicts context words from the target words [25]. Statistically, CBOW smooths over a lot of the distributional information because it treats the entire context as one observation, which works better for larger datasets. However, skip-gram treats each context-target pair as a new observation, which tends to perform better with smaller corpora [25]. Given the size of our corpora, we opt to use skip-gram as our baseline distributional semantic model in the present study.

In the biomedical domain, Pakhomov's work [26] indicates the word2vec representations trained on a clinical corpus of text is able to capture the relationship between biomedical terms. However, this study only utilized the default hyperparameters, such as embedding dimension, for training the word2vec representations. The work of Chiu et al modifies the hyperparameters of word2vec and finds that the performance can be significantly improved in the biomedical domain by hyperparameter tuning [27]. In both Pakhomov's and Chiu's studies, the word representations are still extracted from the vanilla word2vec models, as Pakhomov and Chui only changed the training corpus or the hyperparameters. While corpus-based measures have proven to be more flexible and extensible than ontology-based measures, the lack of both human curation and access to large representative corpora limit their accuracy.

**Hybrid Methods**

There have been previous efforts to combine ontology-based and corpus-based similarities to better capture semantic similarities; however, no framework currently exists in the biomedical domain for incorporating ontological knowledge into the process of generating word embeddings for semantic similarity. Yu and Dredze [22] introduced a general model for learning word embeddings by incorporating prior information. This group proposed relation constrained loss, which is the average log probability of all relations for each term. By adding the relation constrained loss to the original word2vec loss function, this method could include a word's synonyms from WordNet in the word embeddings. The generated word embeddings, produced from the joint objective function and trained on a general corpus, outperformed the baseline word embeddings in three tasks: language modeling, measuring word similarity, and predicting human judgment on word pairs [22]. In addition to word2vec, Alsuhaibani et al. extend the objective function of GloVe in a way similar to Yu and Dredze by adding relation constrained loss to incorporate prior knowledge [28]. Compared to these studies, our method uses a different approach to modify the loss function, as we re-weight the objective function of the word2vec model according to ontology-based semantic similarities for each pair of terms. Particularly, in contrast to these methods, our approach does not introduce additional hyperparameters. As a result, our proposed method does not require careful determination of the value of additional hyperparameters, making the training process more straightforward. Xu et al. [20] introduce RC-NET, a combination of two models, R-NET and C-NET, which use different objective functions to capture relational knowledge and categorical knowledge, respectively. They show that RC-NET, trained on a general corpus, outperforms R-NET, C-NET, and the baseline skip-gram model in the word similarity and topic prediction tasks. Faruqui et al. [21] propose a method for augmenting vector space representations of words using relational information from semantic lexicons. The main contribution of their proposed method, retrofitting, is that it is applied as a post-processing step, which allows it to be used on any pre-trained word vectors [21]. They show that using retrofitting as a post-processing step improves performance on a variety of tasks, including word similarity, syntactic relations, synonym-selection, and sentiment analysis [21]. Finally, Pivovarov and Elhadad [29] present a hybrid score that uses a weighted average of ontology measures and corpus measures to calculate semantic similarity. However, their method doesn't incorporate ontological knowledge into the process for generating the word embeddings; instead, it combines the outputted scores to produce a more accurate final semantic similarity score.

Of note, while there are hybrid methods that combine elements of corpus-based and ontology-based methods, in this paper, we present a general framework for incorporating ontological knowledge into the process for generating word embeddings for semantic similarity in the biomedical domain. We created a new objective function by modifying the Sigmoid cross-entropy objective function of the skip-gram model with ontological knowledge from the MeSH-ontology similarity measures to broaden the domain of our model beyond the scope of the corpus. As a result, our proposed model is able to learn word embeddings that encode both contextual information and domain knowledge, thus making it more accurate and extensible than previous methods.

## METHODS

**Utilized Corpora**

In this work, we use the RadCore and MIMIC-III corpora to train the corpus-based component of our proposed model. Assembled at Stanford in 2007, RadCore is a large multi-institutional radiology report corpus for NLP [30]. The reports in the RadCore corpus range from 1995 to 2006 and were de-identified by their source organizations before submission to RadCore. In its entirety, RadCore contains 1,899,482 reports from three major healthcare organizations: Mayo Clinic (812 reports), MD Anderson Cancer Center (5,000 reports), and Medical College of Wisconsin (1,893,670 reports) [30]. Additionally, all of the radiology reports are in free text format and do not contain any metadata about the type and nature of the imaging exams [27]. Medical Information Mart for Intensive Care (MIMIC-III) is a database containing information gathered from patients that were admitted to critical care units at a large hospital [31]. In this study, we use MIMIC-III's gold standard corpus of 2,434 ICU nursing notes that were "gathered simultaneously with the signals, trends, laboratory reports, discharge summaries and other data in the MIMIC-III databases" [32]. The corpus was thoroughly de-identified; all detected instances of Protected Health Information (PHI) were replaced by realistic surrogate data [32]. The final training corpus, which is a combination of the RadCore and MIMIC-III corpora, contains 195,101,383 total words, 145,274 unique words, and 43,232 unique frequent words with at least five occurrences in the corpora.

**Utilized Ontology-based Measures**

One of the challenges with having numerous medical domain ontologies is that they are typically developed independently of each other and rely on different standards, programming languages, and interfaces [8]. Of note, the UMLS framework, developed by NLM, includes over 100 controlled medical ontologies [8]. Among those, Medical Subject Headings (MeSH) is a controlled hierarchical vocabulary developed by NLM and is widely used for cataloging and searching biomedical and health-related information. In addition, MeSH vocabulary size is more manageable in comparison to larger controlled terminologies, such as SNOMED-CT, thus making it a computationally tractable candidate to develop and evaluate the ontology-based and hybrid semantic similarity measures. In this study, we use three state-of-the-art semantic similarity measures on concepts in the MeSH ontology: Wu & Palmer (wup) [33], Leacock & Chodorow (lch) [34], and Al-Mubaid & Nguyen (nam) [24]. These three semantic similarity measures are defined below.

$$sim_{wup}(c_1, c_2) = \frac{2 * depth(LCS)}{depth(c_1) + depth(c_2)}$$

$$sim_{lch}(c_1, c_2) = -log(\frac{shortest\_path(c_1, c_2)}{2 * depth(ontology)})$$

$$sim_{nam}(c_1, c_2) = log((|shortest_{path(c_1,c_2)}| - 1) * \Big(depth(ontology) - depth\big(lcs(c_1, c_2)\big)\Big) + 2)$$

Before using the ontology-based similarities to modify the objective function of the skip-gram model, we first identified the set of words that appear at the intersection of the set of words in the corpus vocabulary and the set of words that exist in the MeSH hierarchical ontology. Using this intersection set, we generate a similarity matrix containing all pairwise similarities of the intersection terms, normalizing each measure to be in the range of 0 to 1. We utilized an established and widely used Similarity Perl package [8] to compute these similarity measures. For each word pair, if more than one set of ontology-based similarity scores is produced, we calculate the median of the similarity scores for the final similarity score. And, if a similarity score doesn't exist as defined by any of the ontology-based similarity measures, we use a placeholder value of $-1$ to denote that we do not modify the objective function for that particular word pair in the training process. It is important to note that not every pair of words has an ontology-based similarity score because a path may not exist between them in an ontology. Thus, in our study, the final similarity matrix contains 4,878 unique words and 11,945,574 pair-wise similarity scores.

**Multi-Ontology Refined Embeddings—MORE**

Multi-Ontology Refined Embeddings (MORE) is a hybrid semantic similarity measure that effectively integrates ontological knowledge and corpus-based contexts in a novel semantic similarity measure. The ontology-based similarity measures are used to modify the objective function of the word2vec skip-gram model. MORE uses a mathematical framework for vector representation refinement that is extensible in that any number of established and reliable ontology-based measures can be incorporated into the existing framework collection, allowing the model to maximize our measure's utility in a broad domain (see Figure 1).

Traditionally, the objective function of the skip-gram model is a full softmax function. However, the specific implementation of the skip-gram model used for this project relies on a simplified variant of Noise Contrastive Estimation (NCE) [35] that trains faster and results in better vector representations for frequent words, compared to the full softmax function [25]. The loss function, $L_\theta$, is the average Sigmoid cross-entropy loss, which incorporates both the loss computed from the context words, $L_{POS}$, and the loss calculated from the negatively sampled words, $L_{NEG}$, over the batch size:

$$L_\theta = \frac{\Sigma L_{POS} + \Sigma L_{NEG}}{BatchSize}$$

$$L_{POS} = -log(S(logit(w_C|w_I)))$$

$$L_{NEG} = -log(1 - S(logit(w_{NEG}|w_I)))$$

where $S$ is the Sigmoid function, $w_I$ is the input word, $w_C$ is a context word, $w_{NEG}$ is a negatively sampled word, and $logit(w_i|w_I)$ is the log odds of the conditional probability of the label word ($w_C$ or $w_{NEG}$) given the input word, as predicted by the model.

In computing the loss for the context words and negatively sampled words, we modify the binary labels used in the traditional cross-entropy loss function to incorporate the ontology similarities. For the context words, rather than multiplying the negative log of the sigmoid of the model output by one, we multiply it by the average of 1 and the ontology similarity score.

$$L_{POS*} = \frac{1 + sim_{ont}(w_C, w_I)}{2} * -log(S(logit(w_C|w_I)))$$

Similarly, for the negatively sampled words, rather than multiplying the negative log of the sigmoid of the model output by one, we multiply it by one minus the average of zero and the ontology similarity score.

$$L_{NEG*} = (1 - \frac{sim_{ont}(w_{NEG}, w_I)}{2}) * -log(1 - S(logit(w_{NEG}|w_I)))$$

By averaging the binary labels (i.e., 1 and 0) with the similarity scores outputted by the model, the loss function is adjusted to incorporate relational knowledge from the ontologies. For instance, in the case of computing the loss for context words ($L_{POS*}$), if the word pair has a high ontology similarity score, the loss will be higher. In order to minimize loss, the network will adjust the weights in the direction suggested by the ontological knowledge, encouraging the model to output higher probabilities for word pairs with high ontology similarity scores and lower probabilities for word pairs with low ontology similarity scores.

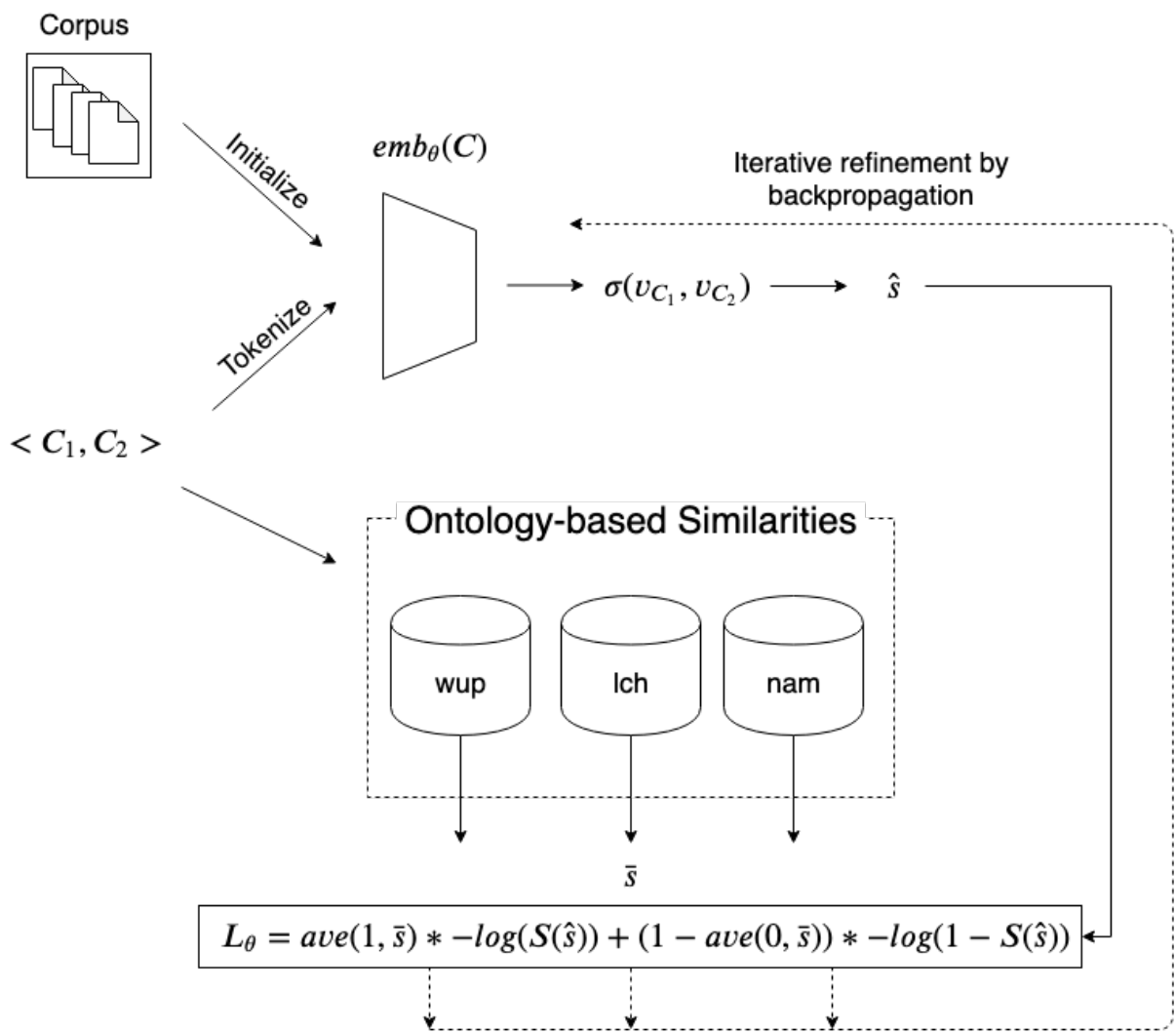

Figure 1: Overview of Multi-Ontology Refined Embeddings (MORE) framework. In training, the similarity between two concepts $(C_1, C_2)$ is measured in different ways: by cosine similarity $\sigma(v_{C_1}, vC_2)$ in a vector space, which gives the skip-gram model's output for the word pair $\hat{s}$, and through ontology-based similarity scores $\bar{s}$. $\bar{s}$ is the median of different ontology-based similarities. The network optimizes the parameters of the skip-gram model by minimizing the modified loss function and backpropagating the loss to refine the embedding layer. The *semantic similarity scores are computed as the cosine similarity of the resulting word embeddings.*

**Evaluation**

In 2007, Pedersen et al. [1] introduced a test set of word pairs for the evaluation of measures of semantic similarity and relatedness in the biomedical domain. They collected 120 concept pairs in the biomedical domain and asked physicians and medical coders to score their similarities. By only selecting the pairs whose inter-rater agreement was high, they curated a reliable test set with 30 concept pairs. Since the introduction of this dataset, it has become the "de facto evaluation standard" and benchmark in the biomedical domain [2] and has been used in multiple studies in this domain for evaluating various semantic similarity measures [2, 8, 9, 24, 29].

These 30 concept pairs of medical terms (See Table 1) were scored by multiple physicians and medical coders on a 4-point scale, according to their relatedness: "practically synonymous (4.0), related (3.0), marginally related (2.0), and unrelated (1.0)" [1]. The average correlation between physicians was 0.68, the average correlation between medical coders was 0.78, and the correlation across groups was 0.85 [1]. In this study, term pair 5, "Delusion — Schizophrenia", has been excluded from the final evaluation dataset because one of the terms did not appear a minimum of five times in our combined corpora. Accordingly, the resulting test set consists of 29 of the 30 original pairs. To evaluate the different methods, we calculated the correlation between the similarity scores outputted by the methods and the physician/medical coder similarity scores. Considering this dataset is relatively small, we also evaluated our model in a larger dataset, which includes 449 concept pairs [26]. The similarity score of each pair of concepts in this dataset was calculated by taking the average of four medical residents' similarity ratings. According to the dataset curation team, 117 of the original 566 concept pairs were discarded from the final dataset because the concepts could not be found in PubMed Central (PMC), a corpus of clinical notes from the Fairview Health System between 2010 and 2014, Wikipedia corpus [26, 37].

In our dataset, there are many biomedical multi-word terms, such as 'Congestive heart failure.' There are two conventional approaches to determine the representations for multi-word terms. The first is constructing a new concept vector directly, and the other is based on the summation/average of component word vectors. Previous research indicates that none of the two aggregation approaches is better than the other one at a statistically significant level [37]. Considering the comparable performance of these two methods, constructing new concept vectors will increase the vocabulary size and, therefore, adds to the computation time and complexity. Thus, in this study, we adopted the second approach and calculated the average of the representation vectors of all words in a multi-word term. We use the average vector as the representation for that multi-word term, and that does not expand the vocabulary size. For each measure of similarity, we calculated the correlation coefficients of the results with human expert ratings and tested the statistical significance of the correlations using a t-test. The null hypothesis in this statistical test is that the correlation coefficient is zero. Of note, for the first dataset, in addition to comparing our model to the method used separately by physicians and medical coders, we mixed the scores of physicians and medical coders and used the merged 58 concept pairs to calculate the combined correlation coefficient.

Table 1: First 10 pairs of evaluation dataset [1].

| Concept 1 | Concept 2 | Physician | Medical Coder |
|---|---|---|---|
| Renal failure | Kidney failure | 4.0000 | 4.0000 |
| Heart | Myocardium | 3.3333 | 3.0000 |
| Stroke | Infarct | 3.0000 | 2.7778 |
| Abortion | Miscarriage | 3.0000 | 3.3333 |
| Delusion | Schizophrenia | 3.0000 | 2.2222 |
| Congestive heart failure | Pulmonary edema | 3.0000 | 1.4444 |
| Metastasis | Adenocarcinoma | 2.6667 | 1.7778 |
| Calcification | Stenosis | 2.6667 | 2.0000 |
| Diarrhea | Stomach cramps | 2.3333 | 1.3333 |
| Mitral stenosis | Atrial fibrillation | 2.3333 | 1.3333 |

## RESULTS

In this section, we compare the proposed model against three established ontology similarity measures and the baseline skip-gram model. The correlation values for the ontology-based measures are extracted from McInnes et al. [8]. The goal of these experiments is to demonstrate the value of using the MORE framework to learn semantic embeddings with information from ontology similarity measures. In each experiment, we compare the baseline embeddings trained with skip-gram against the embeddings trained using the MORE framework. We quantify the evaluation task of measuring semantic similarity utilizing the correlation between the similarity scores generated by the embeddings and the similarity scores produced by expert human raters.

In training the baseline skip-gram model and the proposed model, we used the following default parameters of the TensorFlow implementation of the skip-gram model: embedding size of 300, window size of 10, minimum word count of 5, and a subsampling threshold of 0.001. To expedite the training process, we used a learning rate of 0.3 and a batch size of 1,024. We trained each model for 10 epochs at a time, warm starting each model with the previous model as a checkpoint, for a total of 150 epochs. Table 2 shows a comparison of the best results achieved by all of the models and ontology measures for the 29 concept pairs. Table 3 indicates the comparison of different measures for the large dataset.

Table 2: Similarity correlations of ontology-based measures, skip-gram baseline model, and MORE, with semantic similarity measures from each group of raters and their associate 95% confidence intervals (COIs), for our first test set with 29 biomedical concept pairs.

| Measure | Correlation with Physicians (95% COI) | p-value | Correlation with Coder (95% COI) | p-value | Correlation – Combined (95% COI) | p-value |
|---|---|---|---|---|---|---|
| lch | 0.544 (0.190 - 0.773) | 0.00494** | 0.580 (0.240 - 0.793) | 0.0023694** | 0.541 (0.309 - 0.712) | 4.944e-05**** |
| wup | 0.463 (0.083 - 0.725) | 0.01985* | 0.430 (0.042 - 0.705) | 0.0325* | 0.430 (0.172 - 0.633) | 0.0018355** |
| nam | 0.556 (0.206 - 0.780) | 0.0038963** | **0.613 (0.288 - 0.811)** | 0.001121** | 0.563 (0.337 - 0.727) | 2.11e-05**** |
| skip-gram | 0.668 (0.392 - 0.833) | 0.0001041*** | 0.582 (0.266 - 0.784) | 0.001173** | 0.603 (0.404 - 0.747) | 8.855e-07**** |
| MORE | **0.707 (0.454 - 0.855)** | 2.57e-05***** | 0.604 (0.298 - 0.797) | 0.00067072*** | **0.633 (0.444 - 0.768)** | 1.682e-07**** |

Table 3: Similarity correlations of ontology-based measures, skip-gram baseline model, and MORE, with semantic similarity measures from human raters and their associate 95% confidence intervals (COI), for our second test set with 449 biomedical concept pairs.

| Measure | Correlation with human rater (95% COI) | p-value |
|---|---|---|
| lch | 0.439 (0.334,0.533) | 1.73E-13**** |
| wup | 0.420 (0.313,0.516) | 2.34E-12**** |
| nam | 0.450 (0.346,0.543) | 4.29E-14**** |
| skip-gram | 0.445 (0.344,0.535) | 1.29E-14**** |
| MORE | **0.481 (0.384,0.568)** | < 2.2e-16**** |

## DISCUSSION

We found that, under identical training conditions, MORE consistently outperforms the baseline skip-gram model in terms of correlation with expert-generated similarity scores. For our first dataset with 29 concept pairs, Table 2 illustrates that MORE had a 5.8% higher correlation with the physician similarity scores and a 3.8% higher correlation with the medical coder similarity scores than the baseline skip-gram model. Additionally, MORE had a 27.2% higher correlation with the physician similarity scores than the best ontology similarity measure. Of note, MORE outperformed all except one of the ontology-based measures (nam) in terms of correlation with medical coder similarity scores. The higher correlation between ontology-based measures and medical coders is likely because medical coders were trained to use hierarchical classifications and ontologies to assign similarity scores [1]. Therefore, their performance is more aligned with the structured knowledge in ontologies. The combined correlations in the fourth column of Table 2 show that MORE has the highest combined correlation, which is 5.0% higher than the correlation of the baseline skip-gram model and 12.4% higher than the best ontology measure. For the large evaluation dataset, which includes 449 concept pairs, the MORE model still had the best performance. As shown in Table 3, our MORE model outperforms the baseline skip-gram model by 8.1% and the best ontology-based measure by 6.9% on this dataset. Of note, as shown in Tables 2 and 3, the MORE model achieved the smallest p-value (i.e., the most significant p-value) on both datasets in comparison to the baseline skip-gram model and all ontology-based semantic similarity measures.

As mentioned in the Introduction section, due to the heterogeneity of biomedical concepts, there is no single top-performing corpus-based or ontology-based semantic similarity measure across all applications and domains. However, by modifying the objective function of the skip-gram model with knowledge from the MeSH ontology and multiple ontology-based similarity measures, we can generate embeddings from the RadCore and MIMIC-III corpora that incorporate knowledge beyond the scope of the corpora and maximize the measure's utility in a broad domain. MORE outperforms the baseline skip-gram model in every case, as well as the ontology similarity measures in most cases. As a result, we have demonstrated that the embeddings generated using the MORE framework are more effective at capturing semantic similarity for biomedical concepts, in a broader domain, than any of MORE's individual components.

Despite MORE's promising performance in our evaluation, we recognize that this study has several limitations. First, aside from the increased learning rate and batch size used to expedite training, we used the default training parameters, as suggested by the TensorFlow implementation of the skip-gram model, to train both the baseline skip-gram model and the proposed model. While these parameters have been optimized for training the baseline model, we did not experiment with tuning hyperparameters to optimize the training of the proposed model. However, this suggests that, under equal but potentially sub-optimal training conditions, MORE outperforms the baseline skip-gram model. Second, we have only incorporated three ontology similarity measures (lch, wup, and nam) from one ontology (MeSH) into our novel framework. With a broader range of similarity measures and more ontologies, such as SNOMED-CT, it's possible that MORE could generate embeddings that are more generalizable and accurate than those produced by the present work. Finally, in this study, we only evaluate the quality of the generated word embeddings with a semantic similarity task on two relatively small datasets.

To address these limitations, in future work, we plan to experiment by tuning different training parameters (e.g., learning rate, number of training epochs, and batch size). Furthermore, we plan to extend the model by incorporating more ontologies, such as SNOMED-CT, and other ontology-based similarity measures. Finally, we expect that the proposed framework has further implications beyond semantic similarity. Accordingly, in future work, we plan to evaluate the quality of the MORE embeddings on other extrinsic semantic tasks, such as analogical reasoning, text classification, synonym-selection, and topic modeling.

## CONCLUSION

Learning high-quality word embeddings for semantic similarity in the biomedical domain is valuable for improving the statistical power of NLP analyses, thus making it easier to identify associations between conditions and clinical outcomes in health records and improve information retrieval from scientific journals and clinical reports. To address existing limitations of biomedical semantic similarity measures, we propose a new modified objective function that incorporates domain knowledge into the process for generating word embeddings. In this paper, we presented a novel framework for integrating knowledge from biomedical ontologies into an existing distributional semantic model (i.e., skip-gram) to improve both the flexibility and accuracy of the learned word embeddings. Our implementation is based on the official TensorFlow implementation of word2vec, and we have made it available for public use. We demonstrate that MORE generally outperforms the baseline skip-gram model, as well as the individual ontology-based similarity measures, in computing semantic similarity scores for biomedical word pairs using a benchmark evaluation dataset.

## ACKNOWLEDGMENTS

The authors would like to thank Lamar Moss and Jason Wei for their help with editing the manuscript. The authors also would like to thank Daniel Rubin, Chuck Kahn, Kevin McEnery and Brad Erickson for their help compiling the RadCore database and Daniel Rubin for providing access to the database.

## COMPETING INTERESTS

None Declared.

## FUNDING

This research was supported in part by grants from the National Library of Medicine (R01LM012837) and the National Cancer Institute (R01CA249758).

# REFERENCES

1   Pedersen T, Pakhomov SVS, Patwardhan S, *et al.* Measures of semantic similarity and relatedness in the biomedical domain. *Journal of Biomedical Informatics* 2007;**40**:288–99. doi:10.1016/j.jbi.2006.06.004

2   Sánchez D, Batet M. Semantic similarity estimation in the biomedical domain: An ontology-based information-theoretic perspective. *Journal of Biomedical Informatics* 2011;**44**:749–59. doi:10.1016/j.jbi.2011.03.013

3   Tan WK, Hassanpour S, Heagerty PJ, *et al.* Comparison of Natural Language Processing Rules-based and Machine-learning Systems to Identify Lumbar Spine Imaging Findings Related to Low Back Pain. *Academic Radiology* 2018;**25**:1422–32. doi:10.1016/j.acra.2018.03.008

4   Hassanpour S, Langlotz CP. Information extraction from multi-institutional radiology reports. *Artificial Intelligence in Medicine* 2016;**66**:29–39. doi:10.1016/j.artmed.2015.09.007

5   Huhdanpaa HT, Tan WK, Rundell SD, et al. Using Natural Language Processing of Free-Text Radiology Reports to Identify Type 1 Modic Endplate Changes. J Digit Imaging 2018;31:84–90. doi:10.1007/s10278-017-0013-3

6   Hassanpour S, O'Connor MJ, Das AK. Evaluation of semantic-based information retrieval methods in the autism phenotype domain. *AMIA Annu Symp Proc* 2011;**2011**:569–77.

7   Hassanpour S, Bay G, Langlotz CP. Characterization of Change and Significance for Clinical Findings in Radiology Reports Through Natural Language Processing. *J Digit Imaging* 2017;**30**:314–22. doi:10.1007/s10278-016-9931-8

8   McInnes BT, Pedersen T, Pakhomov SVS. UMLS-Interface and UMLS-Similarity : open source software for measuring paths and semantic similarity. *AMIA Annu Symp Proc* 2009;**2009**:431–5.

9   Batet M, Sánchez D, Valls A. An ontology-based measure to compute semantic similarity in biomedicine. *J Biomed Inform* 2011;**44**:118–25. doi:10.1016/j.jbi.2010.09.002

10  Kumar N, Tafe LJ, Higgins JH, *et al.* Identifying Associations between Somatic Mutations and Clinicopathologic Findings in Lung Cancer Pathology Reports. *Methods Inf Med* 2018;**57**:63–73. doi:10.3414/ME17-01-0039

11  Pesquita C, Faria D, Falcão AO, *et al.* Semantic Similarity in Biomedical Ontologies. *PLoS Comput Biol* 2009;**5**:e1000443. doi:10.1371/journal.pcbi.1000443

12  Rada R, Mili H, Bicknell E, *et al.* Development and application of a metric on semantic nets. *IEEE Trans Syst, Man, Cybern* 1989;**19**:17–30. doi:10.1109/21.24528

13  Mikolov T, Sutskever I, Chen K, *et al.* Distributed Representations of Words and Phrases and their Compositionality. ;:9.

14  De Vine L, Zuccon G, Koopman B, *et al.* Medical Semantic Similarity with a Neural Language Model. In: *Proceedings of the 23rd ACM International Conference on Conference on Information and Knowledge Management - CIKM '14*. Shanghai, China: : ACM Press 2014. 1819–22. doi:10.1145/2661829.2661974

15  Pyysalo S, Ginter F, Moen H, *et al.* Distributional Semantics Resources for Biomedical Text Processing. ;:5.

16  Th M, Sahu S, Anand A. Evaluating distributed word representations for capturing semantics of biomedical concepts. In: *Proceedings of BioNLP 15*. Beijing, China: : Association for Computational Linguistics 2015. 158–63. doi:10.18653/v1/W15-3820

17  Resnik P. Using Information Content to Evaluate Semantic Similarity in a Taxonomy. *arXiv:cmp-lg/9511007* Published Online First: 29 November 1995.http://arxiv.org/abs/cmp-lg/9511007 (accessed 5 Apr 2020).

18  Jiang JJ, Conrath DW. Semantic Similarity Based on Corpus Statistics and Lexical Taxonomy. *arXiv:cmp-lg/9709008* Published Online First: 20 September 1997.http://arxiv.org/abs/cmp-lg/9709008 (accessed 5 Apr 2020).

19  Lin D. An Information-Theoretic Definition of Similarity. ;:9.

20  Xu C, Bai Y, Bian J, *et al.* RC-NET: A General Framework for Incorporating Knowledge into Word Representations. In: *Proceedings of the 23rd ACM International Conference on Conference on Information and Knowledge Management - CIKM '14*. Shanghai, China: : ACM Press 2014. 1219–28. doi:10.1145/2661829.2662038

21  Faruqui M, Dodge J, Jauhar SK, *et al.* Retrofitting Word Vectors to Semantic Lexicons. *arXiv:14114166 [cs]* Published Online First: 22 March 2015.http://arxiv.org/abs/1411.4166 (accessed 5 Apr 2020).

22  Yu M, Dredze M. Improving Lexical Embeddings with Semantic Knowledge. In: *Proceedings of the 52nd Annual Meeting of the Association for Computational Linguistics (Volume 2: Short Papers)*. Baltimore, Maryland: : Association for Computational Linguistics 2014. 545–50. doi:10.3115/v1/P14-2089

23  Bian J, Gao B, Liu T-Y. Knowledge-Powered Deep Learning for Word Embedding. In: Calders T, Esposito F, Hüllermeier E, *et al.*, eds. *Machine Learning and Knowledge Discovery in Databases*. Berlin, Heidelberg: : Springer Berlin Heidelberg 2014. 132–48. doi:10.1007/978-3-662-44848-9_9

24  Al-Mubaid H, Nguyen HA. A Cluster-Based Approach for Semantic Similarity in the Biomedical Domain. In: *2006 International Conference of the IEEE Engineering in Medicine and Biology Society*. New York, NY: : IEEE 2006. 2713–7. doi:10.1109/IEMBS.2006.259235

25  Mikolov T, Chen K, Corrado G, *et al.* Efficient Estimation of Word Representations in Vector Space. *arXiv:13013781 [cs]* Published Online First: 6 September 2013.http://arxiv.org/abs/1301.3781 (accessed 5 Apr 2020).

26  Pakhomov SVS, Finley G, McEwan R, *et al.* Corpus domain effects on distributional semantic modeling of medical terms. *Bioinformatics* 2016;:btw529. doi:10.1093/bioinformatics/btw529

27  Chiu B, Crichton G, Korhonen A, *et al.* How to Train good Word Embeddings for Biomedical NLP. In: *Proceedings of the 15th Workshop on Biomedical Natural Language Processing*. Berlin, Germany: : Association for Computational Linguistics 2016. 166–74. doi:10.18653/v1/W16-2922

28  Alsuhaibani M, Bollegala D, Maehara T, *et al.* Jointly learning word embeddings using a corpus and a knowledge base. *PLoS ONE* 2018;**13**:e0193094. doi:10.1371/journal.pone.0193094

29  Pivovarov R, Elhadad N. A hybrid knowledge-based and data-driven approach to identifying semantically similar concepts. *Journal of Biomedical Informatics* 2012;**45**:471–81. doi:10.1016/j.jbi.2012.01.002

30  Hassanpour S, Langlotz CP. Unsupervised Topic Modeling in a Large Free Text Radiology Report Repository. *J Digit Imaging* 2016;**29**:59–62. doi:10.1007/s10278-015-9823-3

31  Johnson AEW, Pollard TJ, Shen L, *et al.* MIMIC-III, a freely accessible critical care database. *Sci Data* 2016;**3**:160035. doi:10.1038/sdata.2016.35

32  Goldberger AL, Amaral LAN, Glass L, *et al.* PhysioBank, PhysioToolkit, and PhysioNet: Components of a New Research Resource for Complex Physiologic Signals. *Circulation* 2000;**101**. doi:10.1161/01.CIR.101.23.e215

33  Wu Z, Palmer M. Verbs semantics and lexical selection. In: *Proceedings of the 32nd annual meeting on Association for Computational Linguistics  -*. Las Cruces, New Mexico: : Association for Computational Linguistics 1994. 133–8. doi:10.3115/981732.981751

34  Fellbaum C, editor. Combining Local Context and WordNet Similarity for Word Sense Identification. In: *WordNet*. The MIT Press 1998. doi:10.7551/mitpress/7287.003.0018

35  Gutmann MU, Hyv A. Noise-contrastive estimation of unnormalized statistical models, with applications to natural image statistics. ;:55.

36  Pakhomov S, Serguei. Semantic Relatedness and Similarity Reference Standards for Medical Terms. *Retrieved from the Data Repository for the University of Minnesota* Published Online First: 3 May 2018. doi:https://doi.org/10.13020/D6CX04

37  Henry S, Cuffy C, McInnes BT. Vector representations of multi-word terms for semantic relatedness. *Journal of Biomedical Informatics* 2018;**77**:111–9. doi:10.1016/j.jbi.2017.12.006